\documentclass{article}

\usepackage[final,main]{neurips_2026}
\usepackage[utf8]{inputenc}
\usepackage[T1]{fontenc}
\usepackage{hyperref}
\usepackage{url}
\usepackage{booktabs}
\usepackage{amsfonts}
\usepackage{nicefrac}
\usepackage{microtype}
\usepackage{xcolor}
\usepackage{graphicx}
\usepackage{algorithm}
\usepackage{algorithmic}
\usepackage{amssymb}
\usepackage{amsmath}
\usepackage{multirow}
\everydisplay{\small}
\newtheorem{theorem}{Theorem}

\newtheorem{definition}{Definition}
\newtheorem{remark}{Remark}
\usepackage{placeins}
\usepackage{float}
\title{When Scores Conflict with Preferences: \\Pre-Alignment Diagnosis and Adaptive Correction of Constraint Drift in DPO}

\author{%
  Ping Liu \\
  Department of Computer Science \\
  and Engineering \\
  University of Nevada, Reno \\
  \texttt{pino.pingliu@gmail.com} \\
  \And
  Yan Yan\thanks{Corresponding author.} \\
  School of Electrical Engineering \\
  and Computer Science \\
  Washington State University \\
  \texttt{yan.yan1@wsu.edu} \\
}

\begin{document}

\maketitle

\begin{abstract}
Direct Preference Optimization (DPO) is increasingly used in settings where pairwise preferences coexist with pointwise criteria such as response length, answer correctness, or safety scores.
When the two disagree, standard DPO treats pairs whose preferred output scores worse on the pointwise criterion (conflict pairs) in the same way as the remaining (matching) pairs.
We characterize a systematic failure mode, constraint drift: for sufficiently small updates, conflict pairs tend to increase constraint cost while matching pairs tend to repair it, and the net change depends on their balance.
This balance yields a closed-form pre-alignment diagnostic, $\rho^{*}_\mathrm{eq}$, computable from dataset statistics, that indicates whether a dataset is self-correcting or requires intervention.
Building on this diagnostic, we propose Conflict-aware DPO (CoDPO), which down-weights conflict pairs with a strength set from $\rho^{*}_\mathrm{eq}$ rather than tuned, and adds an adaptive margin shift that addresses residual drift during training.
Empirically, drift grows monotonically with conflict exposure on the conflict-dominated domains, and the intervention decision given by $\rho^{*}_\mathrm{eq}$ matches the observed drift on every domain, including one that it identifies as self-correcting.
On all four domains of the main comparison, CoDPO reduces drift while largely preserving the preference learning of vanilla DPO.
Compared with hard filtering and recent constrained and multi-objective baselines, CoDPO also achieves a more favorable drift-preference trade-off.
\end{abstract}

\section{Introduction}
\label{sec:intro}
Generative models are increasingly aligned with human objectives through preference optimization.
Preference optimization methods such as RLHF~\citep{stiennon2020learning_nips2020,ouyang2022training_nips2022,bai2022training_arxiv2022} and DPO~\citep{rafailov2023direct_nips2023} have substantially improved helpfulness, reasoning quality, and instruction following~\citep{azar2024general_aistas2024,ethayarajh2024kto_arxiv2024,meng2024simpo_nips2024,guo2025deepseek_arxiv2025} by learning from pairwise comparisons between candidate outputs.
A response preferred in such a comparison may nevertheless score poorly on a pointwise criterion, such as correctness or safety, that assesses each response on its own~\citep{dai2023safe_arxiv2023,ji2025pku_acl2025}.
Following PSDPO~\citep{meng2026psdpo_mm2026}, we call this coexistence of pairwise preferences and pointwise criteria heterogeneous supervision.

\begin{figure}[t]
\centering
\includegraphics[width=\textwidth]{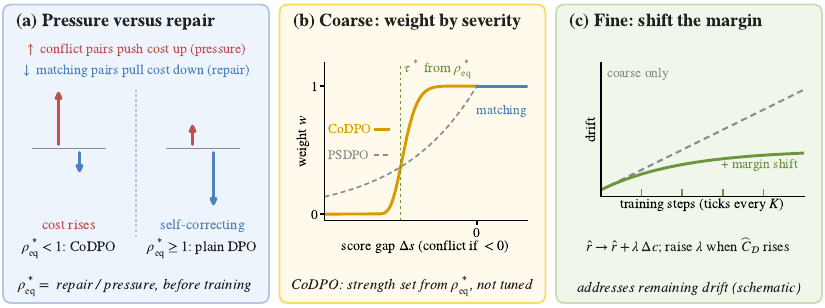}
\caption{Overview of CoDPO.
(a) Under DPO, conflict pairs push the constraint cost up and matching pairs pull it down; $\rho^*_\mathrm{eq}$, computed before training, compares the two.
If the push dominates ($\rho^*_\mathrm{eq} < 1$), CoDPO intervenes; otherwise the data are self-correcting and plain DPO is used.
(b) Conflict pairs are down-weighted by severity, with $\tau^*$ set from $\rho^*_\mathrm{eq}$.
(c) During training, the DPO margin is shifted, more strongly when the constraint-risk estimate rises.
Curves in (b) and (c) are schematic.}
\label{fig:teaser}
\end{figure}

Under heterogeneous supervision, standard preference optimization treats all training pairs uniformly, implicitly assuming that the preferred output also performs well on the pointwise criterion.
On the domains we study (Table~\ref{tab:domains}), the preferred output scores worse on the pointwise criterion in a substantial fraction of pairs: where the criterion is response length, the preferred response is the longer one in 57\% of HelpSteer2~\citep{wang2024helpsteer_nips2024} pairs, and where it is correctness, the preferred response is the incorrect one in 71\% of GSM8K~\citep{cobbe2021training_arxiv2021} pairs and 58\% of ARC~\citep{clark2018think_arxiv2018} pairs.\footnote{GSM8K and ARC pairs are formed from model-generated responses (see the appendix).}
Such pairs, conflict pairs in the terminology of PSDPO~\citep{meng2026psdpo_mm2026}, occur at comparable rates in the other conflict-dominated domains we study; the remaining pairs are matching pairs.
Conflict pairs tend to push preference optimization toward outputs that win the pairwise comparison but carry a higher cost on the pointwise criterion, a degradation that PSDPO~\citep{meng2026psdpo_mm2026} identifies as semantic drift in text-to-video alignment; we call its counterpart under a general pointwise criterion constraint drift.

Existing interventions against constraint drift differ in how they trade the pointwise criterion off against the preference.
Constrained RLHF, constrained DPO~\citep{dai2023safe_arxiv2023,wachi2024stepwise_nips2024,du2025primal_arxiv2025,asadi2025c2_arxiv2025,kim2025safedpo_iclr2026}, and multi-objective methods~\citep{rame2023rewarded_nips2023,zhou2024beyond_acl2024} set this trade-off through a weight that is either fixed by the practitioner or, in Lagrangian methods, adjusted during training according to whether the estimated average cost exceeds a target.
Hard filtering, akin to preference-data selection~\citep{deng2025less_nips2025}, removes conflict pairs, which can discard much of the data when conflicts are frequent, while per-pair weighting~\citep{penguni_iclr2026,meng2026psdpo_mm2026} attenuates them.
In each case, the strength of the intervention is chosen by the practitioner, which leaves two questions open for a given dataset: whether it needs intervention at all, ideally known in advance, and if so, how strong the intervention should be, both before and during training.

We address both questions from a theoretical account of constraint drift under DPO.
We show that, for sufficiently small updates, conflict pairs and matching pairs push the constraint in opposite directions: conflict pairs raise the expected constraint cost, while matching pairs repair it.
Drift therefore depends not on the frequency of conflicts alone but on the balance between conflict-side pressure and matching-side repair.
Under a first-order approximation, this balance can be estimated from dataset statistics alone, before any alignment step is taken.
The estimate yields a closed-form diagnostic, $\rho^*_\mathrm{eq}$, that indicates whether the dataset is self-correcting or needs intervention (Figure~\ref{fig:teaser}a).

When intervention is needed, our control method, Conflict-aware DPO (CoDPO), acts at two levels within a single loss (Figure~\ref{fig:teaser}b,c).
At the coarse level, CoDPO down-weights conflict pairs by severity, rather than removing them all, through asymmetric conflict weighting~\citep{meng2026psdpo_mm2026}.
Instead of tuning the weighting strength, CoDPO solves for it from dataset statistics through $\rho^*_\mathrm{eq}$, so that conflict-side pressure and matching-side repair balance to first order.
The diagnostic is computed at initialization, and the balance can shift as the policy moves away from the reference model during training, so some drift can remain.
The fine level addresses this residual drift by shifting the DPO margin of each pair in proportion to its constraint-cost difference, and strengthens the shift when the drift measured during training rises.
Together, the diagnostic tells the practitioner whether a dataset needs intervention, and CoDPO determines how strong that intervention should be.

We test the diagnostic and CoDPO on five conflict-dominated domains (HelpSteer2, GSM8K, ARC, Math-Calc, TL;DR) and a self-correcting control (PKU-SafeRLHF~\citep{ji2025pku_acl2025}).
On these domains, drift grows monotonically with conflict exposure except on the control, where it stays negligible, and $\rho^*_\mathrm{eq}$ assigns the correct intervention regime on all six.
On the four domains of the main comparison, all of which call for intervention, CoDPO reduces drift while largely preserving the preference learning achieved by vanilla DPO.
Compared with hard filtering and recent constrained and multi-objective baselines, CoDPO also achieves a more favorable drift-preference trade-off.

In summary, our contributions are as follows:
\begin{itemize}
\item We show that, for small updates, the change in constraint cost at each DPO step is a balance between conflict-side pressure and matching-side repair, and derive from it a closed-form diagnostic, $\rho^*_\mathrm{eq}$, indicating before training whether a dataset needs intervention.

\item We propose CoDPO: its coarse level adopts PSDPO's asymmetric weighting form but applies it only when $\rho^*_\mathrm{eq} < 1$, with the strength set from $\rho^*_\mathrm{eq}$ rather than tuned; its fine level shifts the DPO margin and strengthens the shift when the drift measured during training rises.

\item Empirically, the intervention decision given by $\rho^*_\mathrm{eq}$ matches the observed drift on every domain we evaluate, and in our main comparison CoDPO reduces drift while largely preserving vanilla DPO's preference learning, with a more favorable drift-preference trade-off than the compared baselines.
\end{itemize}

\section{Related Work}
\label{sec:related}
\noindent\textbf{Preference and constrained alignment.}
RLHF~\citep{ouyang2022training_nips2022,stiennon2020learning_nips2020} and DPO~\citep{rafailov2023direct_nips2023} learn from pairwise preferences, and subsequent variants modify the preference objective~\citep{azar2024general_aistas2024,ethayarajh2024kto_arxiv2024,meng2024simpo_nips2024,xiao2024cal_nips2024,chowdhury2024provably_icml2024,wu2024beta_nips2024} or address conflicts among multiple pairwise dimensions~\citep{li2025self_acl2025,xu2026understanding_acl2026,zhangdata_iclr2026,chen2026reward_icml2026}, while keeping the supervision purely pairwise.
Constrained alignment adds a pointwise cost, through a cost model with Lagrangian updates in Safe RLHF~\citep{dai2023safe_arxiv2023} and through recent constrained DPO variants~\citep{wachi2024stepwise_nips2024,zhang2025bifactorial_iclr2025,du2025primal_arxiv2025,asadi2025c2_arxiv2025,kim2025safedpo_iclr2026}.
These methods share our supervision structure, pairwise preferences together with a pointwise criterion, and set the trade-off between the two through a parameter chosen by the practitioner or driven by an aggregate cost estimate.
Our analysis addresses a question that precedes this trade-off: whether a given dataset needs constraint handling at all, and if so, how strong it should be, as determined by the conflict structure of its pairs.

\noindent\textbf{Multi-objective optimization and preference-data selection.}
Multi-objective methods such as PCGrad~\citep{yu2020gradient_nips2020}, CAGrad~\citep{liu2021conflict_nips2021}, Rewarded Soups~\citep{rame2023rewarded_nips2023}, and MODPO~\citep{zhou2024beyond_acl2024} address objective competition through gradient projection, model interpolation, or reward combination, and data-centric methods retain a high-margin subset of the training pairs~\citep{deng2025less_nips2025} or select the informative tokens within a pair~\citep{yang2025selective_emnlp2025}; these methods act on objectives or on general data quality rather than on the conflict between a pairwise preference and a pointwise criterion within a pair.
Closer to our coarse level, Uni-DPO~\citep{penguni_iclr2026} keeps every pair but reweights it by an external score margin and by the implicit reward margin, so that ambiguous and already-learned pairs contribute less.
In text-to-video alignment, PSDPO~\citep{meng2026psdpo_mm2026} weights pairs by conflict direction instead, down-weighting conflict pairs with an asymmetric exponential weight that leaves matching pairs unchanged.

Our coarse level adopts this weighting form and differs in how its strength is set.
PSDPO bounds the gradient contributed by conflict pairs and shows that the bound shrinks with stronger down-weighting, which motivates down-weighting but leaves its strength to be tuned.
Section~\ref{sec:theory} instead tracks the constraint cost, which conflict pairs raise and matching pairs reduce.
Because of this repair, the cost stops rising to first order once the conflict weight falls below a critical value, from which CoDPO sets the strength and decides whether down-weighting is needed.
To limit drift that accumulates during training, PSDPO introduces conflict pairs only in a second training stage, whereas our fine level shifts the DPO margin throughout training and strengthens the shift when the measured drift rises.

\section{Constraint Drift as a Balance of Pressure and Repair}
\label{sec:theory}
\subsection{Setup and Conflict Structure}
\label{sec:setup}

Each training example is a preference pair $(x, y_w, y_l)$, in which $y_w$ is the output preferred by the pairwise annotation and $y_l$ the rejected one.
Let $\pi_\theta(y \mid x)$ be a generative policy initialized at the reference policy, $\pi_{\theta_0} = \pi_\mathrm{ref}$, as is standard for DPO, and let $L_R(\theta) = -\mathbb{E}_\mathcal{D}[\log \sigma(\beta\,\hat{r}_\theta)]$ be the DPO preference loss~\citep{rafailov2023direct_nips2023}, where $\hat{r}_\theta = \log\tfrac{\pi_\theta(y_w|x)}{\pi_\mathrm{ref}(y_w|x)} - \log\tfrac{\pi_\theta(y_l|x)}{\pi_\mathrm{ref}(y_l|x)}$ is the implicit reward margin.
A separate scorer $s: \mathcal{X} \times \mathcal{Y} \to \mathbb{R}$ evaluates each output independently of the pairwise signal, oriented so that larger values indicate better satisfaction of the pointwise criterion; for instance, $s$ decreases with response length on a length-constrained domain~\citep{singhal2024long_colm2024,park2024disentangling_acl2024} and is a reward-model score~\citep{ouyang2022training_nips2022,stiennon2020learning_nips2020} on a safety-constrained domain.
A constraint cost $c(x,y) \ge 0$, strictly decreasing in $s$, defines the constraint risk $C(\theta) = \mathbb{E}_{x,\,y\sim\pi_\theta}[c(x,y)]$.

\begin{definition}[Conflict~\citep{meng2026psdpo_mm2026}]
\label{def:conflict}
For a preference pair $(x,y_w,y_l)$, let $\Delta s := s(x,y_w) - s(x,y_l)$ and $\Delta c := c(x,y_w) - c(x,y_l)$.
A conflict pair satisfies $\Delta s < 0$ (equivalently $\Delta c > 0$), meaning that the preferred output scores worse on the pointwise criterion than the rejected one.
We define the conflict set $\mathcal{D}_\mathrm{conf} := \{(x,y_w,y_l): \Delta s < 0\}$ and the matching set $\mathcal{D}_\mathrm{match} := \{(x,y_w,y_l): \Delta s \ge 0\}$, with conflict rate $p_c := \Pr(\Delta s < 0)$.\footnote{Pairs with $\Delta s = 0$ satisfy $\Delta c = 0$. We place them in $\mathcal{D}_\mathrm{match}$ so that the partition is exhaustive and, under Assumption (vii) of Appendix~B.1, define their composite contribution $|\Delta c|\,\mathcal{I}_\theta$ to be zero; $\mathcal{I}_\theta$ itself need not be evaluated at $\Delta c = 0$.}
\end{definition}

\subsection{Pressure, Repair, and the Diagnostic}
\label{sec:drift}
Following PSDPO~\citep{meng2026psdpo_mm2026}, the conflict-matching partition splits the DPO gradient into contributions from the two subsets,
\[
\nabla L_R(\theta) = (1-p_c)\,g_\mathrm{match} + p_c\,g_\mathrm{conf},
\]
where $g_\mathrm{match} = \mathbb{E}_{\mathcal{D}_\mathrm{match}}[\nabla\ell_R]$, $g_\mathrm{conf} = \mathbb{E}_{\mathcal{D}_\mathrm{conf}}[\nabla\ell_R]$, and the per-sample loss is $\ell_R = -\log\sigma(\beta\hat{r}_\theta)$.
Whereas PSDPO~\citep{meng2026psdpo_mm2026} bounds the norm of the conflict term, we bound its first-order effect on the constraint risk, as in first-order analyses of DPO training dynamics~\citep{razin2025unintentional_iclr2025,ren2025learning_iclr2025}: under regularity conditions on $C$ and the coupling argument of Appendix~B.2, the conflict term can raise the constraint risk to first order,
\[
-\langle \nabla C, g_\mathrm{conf}\rangle \ge \mathbb{E}_{\mathcal{D}_\mathrm{conf}}\!\left[ \beta(1-\sigma(\beta\hat{r}_\theta))\,\Delta c\,\mathcal{I}_\theta \right] \ge 0,
\]
where, for $\Delta c \neq 0$, the geometry term $\mathcal{I}_\theta := (\Delta c)^{-1}\langle \nabla_\theta C_x(\theta), \nabla_\theta \hat{r}_\theta \rangle > 0$, with $C_x(\theta) = \mathbb{E}_{y\sim\pi_\theta}[c(x,y)]$, measures how much the expected cost on prompt $x$ changes, per unit of the cost gap $\Delta c$, when DPO learns the pair; it is named for the inner product between the two gradient directions.
An analogous bound with the opposite sign holds for $g_\mathrm{match}$ and captures the repair provided by matching pairs.

\begin{theorem}[Constraint drift]
\label{thm:drift}
Under the regularity conditions of Appendix~B.1 (proof in Appendix~B.3), gradient descent on $L_R$ satisfies
\begin{equation}\label{eq:drift}
\mathbb{E}[C(\theta_{t+1}) - C(\theta_t)] \ge -\eta(1-p_c)\mu_0 + \eta\,p_c\,\kappa_0 - \tfrac{1}{2}\eta^2 L_C G_R^2,
\end{equation}
where $\eta$ is the step size, $L_C$ is the smoothness constant of $C$, and $G_R$ bounds $\|\nabla\ell_R\|$.
The terms
\[
\kappa_0 := \mathbb{E}_{\mathcal{D}_\mathrm{conf}}\!\left[ \beta(1-\sigma(\beta\hat r_\theta))\,\Delta c\,\mathcal I_\theta \right] > 0,
\qquad
\mu_0 := \mathbb{E}_{\mathcal{D}_\mathrm{match}}\!\left[ \beta(1-\sigma(\beta\hat r_\theta))\,|\Delta c|\,\mathcal I_\theta \right] > 0
\]
aggregate the conflict-side drift pressure and the matching-side repair, respectively.
For sufficiently small $\eta$, whenever $p_c\kappa_0 > (1-p_c)\mu_0$, the lower bound is strictly positive and the expected constraint risk increases.
\end{theorem}

Theorem~\ref{thm:drift} thus expresses each DPO step as a balance between conflict-side pressure and matching-side repair: drift appears when pressure outweighs repair, and we call the reverse case, in which repair at least matches pressure, the self-correcting regime, which Section~\ref{sec:attenuation} makes precise.

\begin{remark}[Pre-alignment diagnostic]
\label{rmk:early_diag}
The drift condition of Theorem~\ref{thm:drift}, $p_c\kappa_0 > (1-p_c)\mu_0$, is equivalent to $(1-p_c)\mu_0/(p_c\kappa_0) < 1$, and this ratio admits a simple data-dependent approximation at the start of alignment.
Both $\mu_0$ and $\kappa_0$ average the product of three factors: the learning signal, the cost difference, and the geometry term.
At initialization $\pi_\theta = \pi_\mathrm{ref}$, so $\hat{r}_\theta = 0$ and the learning signal $\beta(1-\sigma(\beta\hat{r}_\theta))$ takes the same value $\beta/2$ on every pair, which cancels exactly in the ratio.
Assumption (vi) (Appendix~B.1) posits that, per unit of cost gap, learning a conflict pair changes the expected cost by about as much as learning a matching pair; under this assumption, the geometry term cancels approximately as well, and the ratio reduces to the average cost differences $\overline{\Delta c}_\mathrm{conf} := \mathbb{E}_{\mathcal{D}_\mathrm{conf}}[\Delta c]$ and $\overline{|\Delta c|}_\mathrm{match} := \mathbb{E}_{\mathcal{D}_\mathrm{match}}[|\Delta c|]$, giving the closed-form diagnostic
\begin{equation}\label{eq:rho_eq}
\rho^*_\mathrm{eq} = \frac{(1-p_c)\,\overline{|\Delta c|}_\mathrm{match}}{p_c\,\overline{\Delta c}_\mathrm{conf}},
\end{equation}
that is, the total cost reduction across matching pairs divided by the total cost increase across conflict pairs, so $\rho^*_\mathrm{eq}$ depends not only on how often conflicts occur but also on how large the cost differences are.
Under this approximation, $\rho^*_\mathrm{eq} \ge 1$ indicates before alignment that the dataset is self-correcting and needs no intervention, while a value below $1$ calls for intervention; Appendix~B.5 gives the full derivation.
\end{remark}

\subsection{Down-Weighting Conflict Pairs}
\label{sec:attenuation}

Theorem~\ref{thm:drift} suggests keeping matching pairs at full weight and down-weighting conflict pairs, as PSDPO~\citep{meng2026psdpo_mm2026} does in text-to-video alignment; the question we address is how much down-weighting is enough.
We assign each pair a weight $w(\Delta s)$, equal to $1$ on matching pairs and in $[0,1]$ on conflict pairs, and summarize how much of the conflict-side signal is retained by the average conflict weight $\rho := \mathbb{E}_{\mathcal{D}_\mathrm{conf}}[w(\Delta s)]$, so that $\rho = 1$ recovers plain DPO and $\rho = 0$ removes all conflict pairs.
Theorem~\ref{thm:attenuation} answers this question with a critical value $\rho^*$ for the average conflict weight, below which the reduced conflict-side pressure is smaller than the matching-side repair.

\begin{theorem}[Drift attenuation]
\label{thm:attenuation}
Under the conditions of Theorem~\ref{thm:drift}, the modulated loss $L_{R,w}(\theta) = -\mathbb{E}[w(\Delta s)\log\sigma(\beta\hat{r}_\theta)]$ satisfies the upper bound
\[
\mathbb{E}[C(\theta_{t+1}) - C(\theta_t)] \le -\eta(1-p_c)\mu_0 + \eta\,p_c\,\kappa_w + \tfrac{1}{2}\eta^2 L_C G_R^2,
\]
where $\kappa_w := \mathbb{E}_{\mathcal{D}_\mathrm{conf}}[ w(\Delta s)\,\beta(1-\sigma(\beta\hat r_\theta))\,\Delta c\,\mathcal I_\theta ]$ is the modulated counterpart of $\kappa_0$ and satisfies $\kappa_w \le \rho\,\kappa_0$ by Assumption (v) and Chebyshev's sum inequality (Appendix~B.4).
Using $\kappa_w \le \rho\,\kappa_0$, the first-order part of this bound is at most $\eta\,[\,p_c\,\rho\,\kappa_0 - (1-p_c)\,\mu_0\,]$, which is negative when $\rho$ lies below the critical weight
\[
\rho^* = \frac{(1-p_c)\mu_0}{p_c\kappa_0},
\]
the model-dependent threshold at which conflict-side pressure and matching-side repair balance.
For $\rho < \rho^*$ and a step size satisfying the explicit bound given in Appendix~B.4, the expected per-step change in constraint risk is non-positive.
\end{theorem}

Down-weighting leaves the repair from matching pairs unchanged and reduces the pressure from conflict pairs to at most $\rho$ times its original value.
Once $\rho$ falls below $\rho^*$, the reduced pressure is smaller than the repair, so for a sufficiently small step each update is expected to lower the constraint risk or leave it unchanged.
Plain DPO corresponds to $\rho = 1$, so when $\rho^* > 1$ it already satisfies this condition, and the data are self-correcting.

To use this result, CoDPO needs a value of $\rho$ before training starts.
Because $\rho^*$ depends on gradients of the model, CoDPO instead uses its estimate $\rho^*_\mathrm{eq}$ from Remark~\ref{rmk:early_diag}, which needs only the cost gaps in the data.
It sets the average conflict weight to $\rho = \rho^*_\mathrm{eq}$, and keeps every weight at $1$, that is, plain DPO, when $\rho^*_\mathrm{eq} \ge 1$.
Some drift can nevertheless remain, for two reasons: $\rho$ is set at the estimated balance point rather than strictly below $\rho^*$, and $\rho^*$ itself shifts as training proceeds.
Section~\ref{sec:method} implements this coarse level and adds a fine level that addresses this residual drift.

\section{CoDPO: Controlling the Balance Before and During Training}
\label{sec:method}
CoDPO turns this analysis into two levels of control: a coarse level that sets the conflict weights before training (Section~\ref{sec:coarse}) and a fine level that corrects the drift remaining during training (Section~\ref{sec:fine_correction}).

\subsection{Coarse Level: Setting the Weights Before Training}
\label{sec:coarse}
The coarse level assigns per-pair weights whose average conflict weight matches the pre-alignment estimate $\rho^*_\mathrm{eq}$ of Remark~\ref{rmk:early_diag}.\footnote{Two degenerate cases of Eq.~\eqref{eq:rho_eq} are fixed by convention, each the limit of the expression: if $p_c = 0$ there are no conflict pairs and $\rho^*_\mathrm{eq} = +\infty$, leaving the modulation off; if $p_c = 1$ there are no matching pairs and $\rho^*_\mathrm{eq} = 0$, removing conflict pairs.}
We keep matching pairs at full weight and down-weight conflict pairs by severity through the weighting family
\begin{equation}\label{eq:w_exp}
w_{\tau,p}(\Delta s) = \begin{cases} 1, & \Delta s \ge 0, \\ \exp\!\bigl(-(|\Delta s| / \tau)^p\bigr), & \Delta s < 0, \end{cases}
\end{equation}
where $\tau$ sets how strongly conflict pairs are down-weighted overall and $p$ sets how sharply the weight drops from $1$ to near $0$ as the conflict becomes more severe.\footnote{We use $p=8$ in all experiments. As $p \to \infty$ the modulation approaches a hard threshold at $|\Delta s| = \tau$, which retains mild conflicts and removes severe ones; this differs from $\rho \to 0$, which removes conflict pairs regardless of severity.}
This family extends the asymmetric exponential weighting introduced by PSDPO~\citep{meng2026psdpo_mm2026} with the sharpness $p$ and recovers it at $p=1$; in PSDPO, $\tau$ is chosen as a hyperparameter.
In CoDPO, $\tau$ is instead determined by the diagnostic.
For fixed $p$, each $\tau$ yields an average conflict weight $\rho(\tau) := \mathbb{E}_{\mathcal{D}_\mathrm{conf}}[w_{\tau,p}(\Delta s)]$, which increases strictly from $0$ to $1$ as $\tau$ grows from $0$ to $\infty$: a small $\tau$ suppresses nearly all conflict pairs, and a large $\tau$ keeps them at nearly full weight.
CoDPO picks the $\tau$ whose average conflict weight equals the diagnostic,
\begin{equation}\label{eq:calibration}
\rho(\tau^*) = \mathrm{clip}(\rho^*_\mathrm{eq},0,1).
\end{equation}
Because $\rho(\tau)$ is strictly increasing, Eq.~\eqref{eq:calibration} has a unique solution for $0 < \rho^*_\mathrm{eq} < 1$, which we locate by bisection; when $\rho^*_\mathrm{eq} \ge 1$, CoDPO keeps $w_i = 1$ on every pair.
The calibration uses only dataset statistics and requires no gradient computation.

\subsection{Fine Level: Shifting the Margin During Training}
\label{sec:fine_correction}
The coarse weights are calibrated from $\rho^*_\mathrm{eq}$, which assumes the learning signal $\beta(1-\sigma(\beta\hat r_\theta))$ to be the same on every pair.
This holds at initialization, where $\pi_\theta = \pi_\mathrm{ref}$ gives every pair a zero margin $\hat r_\theta$.
As training proceeds, the learning signal comes to differ from pair to pair, so the balance point that $\rho^*_\mathrm{eq}$ approximates moves while the weights stay fixed, and the constraint risk can keep rising; we call this remaining drift residual drift.
Lowering the conflict weights further would discard more of the preference signal that conflict pairs carry, and fixed weights cannot respond to drift as it emerges, so we add a second control that acts during training without changing the weights.

\noindent\textbf{Shifting the margin.}
DPO pushes each pair with strength $\beta(1-\sigma(\beta\hat r_\theta))$, which is set by the pair's margin $\hat r_\theta$: the larger the margin, the more the pair is treated as already learned and the less it is pushed.
Adding an offset to a pair's margin therefore makes DPO update that pair more or less strongly, while leaving its coarse weight $w$ unchanged.
On the $i$-th pair $(x_i, y_{w,i}, y_{l,i})$, DPO moves probability from $y_{l,i}$ toward $y_{w,i}$.
On a conflict pair ($\Delta c_i > 0$), $y_{w,i}$ is the costlier response, so this raises the expected constraint cost; on a matching pair ($\Delta c_i < 0$), it lowers it.
To push conflict pairs less and matching pairs more, we shift each pair's margin in proportion to its cost gap,
\[
\hat{r}_\theta \;\to\; \hat{r}_\theta + \lambda\,\Delta c, \qquad \lambda \ge 0.
\]
A conflict pair has its margin raised and is pushed less, most of all when its conflict is severe, so its effect on the constraint cost weakens while its preference is still learned; a matching pair has its margin lowered and is pushed more, so its repair strengthens.
A single scalar $\lambda$ thus adjusts each pair according to its own cost gap and, unlike the fixed margin offsets of ODPO~\citep{amini2024odpo_acl2024}, R-DPO~\citep{park2024disentangling_acl2024}, and SafeDPO~\citep{kim2025safedpo_iclr2026}, can be changed during training.

\noindent\textbf{Implementing the shift.}
The margin shift $\hat r_\theta \to \hat r_\theta + \lambda\Delta c$ needs no change to the DPO loss: standard DPO computes it when the reference is tilted against high-cost outputs,
\begin{equation}\label{eq:tilted_ref}
\tilde{\pi}_\mathrm{ref}(y \mid x) \propto \pi_\mathrm{ref}(y \mid x)\,\exp\bigl(-\lambda\,c(x,y)\bigr),
\end{equation}
since $\log\tilde{\pi}_\mathrm{ref}(y \mid x) = \log\pi_\mathrm{ref}(y \mid x) - \lambda\,c(x,y) - \log Z(x)$ and the normalizer $Z(x)$, shared by $y_w$ and $y_l$, cancels in the pairwise margin, leaving $\tilde{r}_\theta = \hat{r}_\theta + \lambda\,\Delta c$ (Appendix~C.2).
Running standard DPO with $\tilde{\pi}_\mathrm{ref}$ thus replaces $\ell_R$ by $\ell_\mathrm{tilt} = -\log\sigma(\beta\tilde{r}_\theta)$, and since $\lambda\Delta c$ does not depend on $\theta$, the gradient direction is unchanged and only the argument of $\sigma$ shifts:
\begin{equation}\label{eq:tilted_grad}
\nabla\ell_\mathrm{tilt} = -\beta\Bigl(1-\sigma\bigl(\beta\hat{r}_\theta + \beta\lambda\Delta c\bigr)\Bigr)\,\nabla\hat{r}_\theta.
\end{equation}
The tilt also links the shift to constrained alignment: because the optimal DPO policy is proportional to the reference times $\exp(r/\beta)$, training against $\tilde{\pi}_\mathrm{ref}$ targets the cost-penalized reward $r - \beta\lambda c$, the form used by Lagrangian approaches~\citep{dai2023safe_arxiv2023,du2025primal_arxiv2025}, which typically raise $\lambda$ while an estimated cost exceeds a prescribed threshold; CoDPO instead raises $\lambda$ when the constraint risk measured on the training pairs rises, as described next.

\begin{algorithm}[t]
\footnotesize
\caption{CoDPO training procedure}
\label{alg:codpo}
\begin{algorithmic}[1]
\REQUIRE Preference dataset $\mathcal{D}$, scorer $s$, cost $c$, reference policy $\pi_\mathrm{ref}$, temperature $\beta$, learning rate $\eta$, steps $T$, interval $K$, step fraction $\alpha$
\ENSURE Trained policy $\pi_{\theta^*}$
\STATE Compute $\Delta s_i$ and $\Delta c_i$ for all pairs
\STATE Compute $p_c$ and $\rho^*_\mathrm{eq}$, using the degenerate conventions of \S\ref{sec:coarse} when $p_c \in \{0,1\}$
\IF{$\rho^*_\mathrm{eq} \ge 1$}
  \STATE Set $w_i \leftarrow 1$ for all pairs
\ELSIF{$\rho^*_\mathrm{eq} = 0$}
  \STATE Set $w_i \leftarrow 1$ for matching pairs and $w_i \leftarrow 0$ for conflict pairs
\ELSE
  \STATE Solve Eq.~\eqref{eq:calibration} for $\tau^*$ by bisection and set $w_i$ using Eq.~\eqref{eq:w_exp}
\ENDIF
\STATE Compute $\lambda_\mathrm{tilt}$ using Eq.~\eqref{eq:lambda_tilt}
\STATE Initialize $\lambda \leftarrow \alpha\lambda_\mathrm{tilt}$ and $C_{\mathrm{prev}} \leftarrow \widehat{C}_\mathcal{D}(\theta_0)$
\FOR{$t=1,\ldots,T$}
  \STATE Sample mini-batch $\mathcal{B}$
  \STATE Update $\theta$ using $\nabla_\theta \mathcal{L}_\mathrm{CoDPO}(\theta)$
  \IF{$t \bmod K = 0$}
    \STATE $C_{\mathrm{curr}} \leftarrow \widehat{C}_\mathcal{D}(\theta_t)$
    \IF{$C_{\mathrm{curr}} > C_{\mathrm{prev}}$}
      \STATE $\lambda \leftarrow \min(\lambda+\alpha\lambda_\mathrm{tilt},\lambda_\mathrm{tilt})$
    \ENDIF
    \STATE $C_{\mathrm{prev}} \leftarrow C_{\mathrm{curr}}$
  \ENDIF
\ENDFOR
\end{algorithmic}
\end{algorithm}

\noindent\textbf{Setting the shift strength online.}
How much the constraint cost still rises under the coarse weights depends on how training unfolds, which cannot be known before training, so $\lambda$ is adjusted during training.
Adjusting $\lambda$ requires deciding when to increase it and by how much.
When to increase $\lambda$ is decided by the change in constraint risk, measured every $K$ training steps on the training pairs through the estimate
\begin{equation}\label{eq:chat}
\widehat{C}_\mathcal{D}(\theta) = \frac{1}{|\mathcal{D}|}\sum_i \bigl[ q_{\theta,i}\,c(x_i,y_{w,i}) + (1-q_{\theta,i})\,c(x_i,y_{l,i}) \bigr],
\qquad
q_{\theta,i} = \frac{\pi_\theta(y_{w,i} \mid x_i)}{\pi_\theta(y_{w,i} \mid x_i) + \pi_\theta(y_{l,i} \mid x_i)},
\end{equation}
where $q_{\theta,i}$ is the probability the current policy assigns to $y_{w,i}$ within the pair, a normalized policy probability rather than the Bradley--Terry probability $\sigma(\beta\hat r_\theta)$.
Because $\widehat{C}_\mathcal{D}$ uses only the responses $y_{w,i}$ and $y_{l,i}$ already in the training data, rather than new responses sampled from the policy, it adds no generation cost; it rises when, across pairs, the policy shifts probability toward the costlier response of each pair on balance.
How much to increase it is set by the same first-order analysis as $\rho^*_\mathrm{eq}$, which, under the same approximations and for the unmodulated loss, gives the shift strength that cancels first-order drift (Appendix~C.4),
\begin{equation}\label{eq:lambda_tilt}
\lambda_\mathrm{tilt} = \max\!\left( 0,\; \frac{2\,\mathbb{E}_{\mathcal{D}}[\Delta c]}{\beta\,\mathbb{E}_{\mathcal{D}}[\Delta c^2]} \right).
\end{equation}
Because the coarse weights already remove part of this drift, setting $\lambda = \lambda_\mathrm{tilt}$ outright could overcorrect, so we increase $\lambda$ gradually, in increments of $\alpha\lambda_\mathrm{tilt}$, and cap it at $\lambda_\mathrm{tilt}$.
Concretely, $\lambda$ starts at $\alpha\lambda_\mathrm{tilt}$, so that the shift is active from the first step; whenever $\widehat{C}_\mathcal{D}$ has increased between two consecutive evaluations, which indicates that the residual drift has grown and the current shift is too weak, $\lambda$ is raised by $\alpha\lambda_\mathrm{tilt}$, up to $\lambda_\mathrm{tilt}$, and otherwise it is kept unchanged, since a stable or falling $\widehat{C}_\mathcal{D}$ indicates that the current shift suffices.
Taking the maximum with $0$ in $\lambda_\mathrm{tilt}$ prevents a negative $\lambda$, which would reverse the shift.
It also switches the fine level off on self-correcting data: there $\mathbb{E}_{\mathcal{D}}[\Delta c] \le 0$, which is equivalent to $\rho^*_\mathrm{eq} \ge 1$, so $\lambda_\mathrm{tilt} = 0$ and, with all coarse weights at $1$, CoDPO reduces to vanilla DPO.

\subsection{Combining the Two Levels}
\label{sec:algorithm}
CoDPO combines the coarse weights and the margin shift in a single loss:
\begin{equation}\label{eq:combined_loss}
\mathcal{L}_\mathrm{CoDPO}(\theta) = -\frac{1}{|\mathcal{B}|}\sum_{i \in \mathcal{B}} w_i \cdot \log\sigma\bigl(\beta\,\hat{r}_\theta^{(i)} + \beta\lambda\Delta c_i\bigr),
\end{equation}
where the weights $w_i$ are fixed before training (Section~\ref{sec:coarse}) and the shift $\lambda\Delta c_i$ is adjusted during training (Section~\ref{sec:fine_correction}).
The complete training procedure is summarized in Algorithm~\ref{alg:codpo}.
The offline calibration steps are performed once before training, in $O(|\mathcal{D}|)$ time per bisection iteration.
Each optimization step has the same cost as vanilla DPO because $w_i$, $\Delta c_i$, and $\lambda$ enter the loss only as scalar coefficients.
Additional overhead comes only from evaluating the constraint-risk estimate $\widehat{C}_\mathcal{D}(\theta_t)$ to update $\lambda$, which uses forward passes on the fixed training pairs and no generation, and is needed only once every $K=500$ steps.

\section{Experiments}
\label{sec:experiments}

We evaluate on a suite of domains spanning diverse constraint types (Table~\ref{tab:domains}).
Most are conflict-dominated, where preference and pointwise criterion frequently disagree (HelpSteer2~\citep{wang2024helpsteer_nips2024}, GSM8K~\citep{cobbe2021training_arxiv2021}, ARC~\citep{clark2018think_arxiv2018}, Math-Calc, TL;DR~\citep{stiennon2020learning_nips2020}).
PKU-SafeRLHF~\citep{ji2025pku_acl2025} serves as a self-correcting control where the diagnostic predicts no intervention ($\rho^*_\mathrm{eq} \ge 1$, clipped to 1.0).
Math-Calc is constructed from GSM8K~\citep{cobbe2021training_arxiv2021} with calculation density as the preference signal.
We use the conflict-dominated domains for drift-curve validation and a representative subset (HelpSteer2, GSM8K, ARC, Math-Calc) for full method comparison against baselines.

\begin{table}[t]
\centering
\caption{Experimental domains. $p_c$: conflict rate. $\rho^*_\mathrm{eq}$: pre-alignment diagnostic (clipped to $[0,1]$). Zero-cross.: estimated value of $\rho$ at which drift changes sign in Table~\ref{tab:drift_curves}.}
\label{tab:domains}
\scriptsize
\begin{tabular}{llllccc}
\toprule
Domain & Pairwise preference & Pointwise constraint & Pairs & $p_c$ & $\rho^*_\mathrm{eq}$ & Zero-cross. \\
\midrule
HelpSteer2 & Helpfulness rating & Response length & 7{,}220 & 57\% & 0.71 & $\sim$0.76 \\
GSM8K & Explanation detail & Answer correctness & 14{,}284 & 71\% & 0.40 & $\sim$0.27 \\
ARC & Reward-model score & Answer correctness & 12{,}580 & 58\% & 0.72 & $\sim$0.61 \\
Math-Calc & Calculation density & Answer correctness & 10{,}387 & 56\% & 0.80 & $\sim$0.50 \\
TL;DR & Human preference & Summary length & 92{,}534 & 58\% & 0.59 & $\sim$0.43 \\
PKU-SafeRLHF & Helpfulness & Safety score & 73{,}907 & 2.5\% & 1.0 & --- \\
\bottomrule
\end{tabular}
\end{table}

We use Qwen2.5-7B-Instruct~\citep{yang2024qwen25_arxiv2024} as the default base model with LoRA~\citep{hu2022lora_iclr2022}, DPO with $\beta{=}0.1$, and CoDPO with $p{=}8$, $K{=}500$, $\alpha{=}0.1$ (full hyperparameters in Appendix~A).
For robustness, we additionally test three base models (DeepSeek-R1-Distill-Llama-8B~\citep{guo2025deepseek_arxiv2025}, Mistral-7B-Instruct-v0.3~\citep{jiang2023mistral_arxiv2023}, Qwen3-8B~\citep{yang2025qwen3_arxiv2025}) and three preference losses (IPO~\citep{azar2024general_aistas2024}, SimPO~\citep{meng2024simpo_nips2024}, KTO~\citep{ethayarajh2024kto_arxiv2024}).
We compare against five recent methods: SafeDPO~\citep{kim2025safedpo_iclr2026}, PD-DPO~\citep{du2025primal_arxiv2025}, C2-DPO~\citep{asadi2025c2_arxiv2025}, MODPO~\citep{zhou2024beyond_acl2024}, and Rewarded Soups~\citep{rame2023rewarded_nips2023}, with vanilla DPO ($\rho{=}1$) and hard filtering ($\rho{=}0$) as references.

We evaluate each model on the two candidate outputs in each preference pair.
The expected constraint cost is
\[
\mathbb{E}[c] = P_\theta(y_w \succ y_l \mid x)\,c(x,y_w) + P_\theta(y_l \succ y_w \mid x)\,c(x,y_l),
\]
where the pairwise probabilities are obtained from the policy's likelihoods of the two candidates.
Drift is measured as $\mathbb{E}[c(\theta)]-\mathbb{E}[c(\theta_\mathrm{ref})]$, preference learning as $P_\theta(\mathrm{chosen})=P_\theta(y_w \succ y_l \mid x)$, and generation quality as pairwise LLM-as-judge win rates using JudgeLM-13B~\citep{zhu2023judgelm_iclr2025} with position-debiased scoring.

\subsection{Drift Curves and Diagnostic Validation}
\label{sec:exp_drift}
To test the balance in Theorem~\ref{thm:drift}, we vary the average conflict weight $\rho$ over $\{0, 0.3, 0.5, 0.7, 1\}$ using the modulation family of CoDPO, train a model at each value, and measure drift and $P(\mathrm{chosen})$ on the training pairs (Table~\ref{tab:drift_curves}).
The two ends of this range correspond to familiar settings: $\rho = 0$ removes all conflict pairs, as hard filtering does, and $\rho = 1$ is vanilla DPO.

Table~\ref{tab:drift_curves} shows three patterns that hold in all five conflict-dominated domains.
First, drift increases monotonically with $\rho$.
Second, drift is negative at $\rho = 0$, where only matching pairs drive training, and positive at $\rho = 1$, where conflict pairs keep full weight, consistent with the balance between repair and pressure in Theorem~\ref{thm:drift}.
Third, $P(\mathrm{chosen})$ largely increases with $\rho$, so removing conflict pairs to suppress drift also weakens preference learning.
Hard filtering ($\rho = 0$) keeps only 7\% to 41\% of the gain in $P(\mathrm{chosen})$ that vanilla DPO achieves over the base model on HelpSteer2, GSM8K, ARC, and Math-Calc, and on TL;DR it falls below the base model.
Removing all conflict pairs thus suppresses drift at a large cost in preference learning, which motivates keeping $\rho$ near the value at which drift changes sign rather than at $0$.

This value can be read off Table~\ref{tab:drift_curves} (Zero-cross.\ in Table~\ref{tab:domains}).
Computed from dataset statistics alone, the diagnostic agrees with this behavior: $\rho^*_\mathrm{eq} < 1$ in all five domains, matching the positive drift at $\rho = 1$, and $\rho^*_\mathrm{eq}$ lies within 0.05 to 0.30 of the zero crossing, a gap we attribute to the first-order approximation.
On PKU-SafeRLHF, $\rho^*_\mathrm{eq} \ge 1$ matches its negligible drift.
The diagnostic thus decides whether to intervene and sets the strength of the coarse weights.

\begin{table}[t]
\centering
\caption{Drift curves across six domains. Each row corresponds to a different average conflict weight $\rho$; Base denotes the reference model before alignment. The constraint cost is response length (characters) on HelpSteer2 and TL;DR, binary incorrectness probability on GSM8K, ARC, and Math-Calc, and based on the safety score on PKU-SafeRLHF (Appendix~A).}
\label{tab:drift_curves}
\scriptsize\setlength{\tabcolsep}{2.7pt}
\begin{tabular}{l|ccc|ccc|ccc|ccc|ccc|ccc}
\toprule
 & \multicolumn{3}{c|}{HelpSteer2} & \multicolumn{3}{c|}{GSM8K} & \multicolumn{3}{c|}{ARC} & \multicolumn{3}{c|}{Math-Calc} & \multicolumn{3}{c|}{TL;DR} & \multicolumn{3}{c}{PKU-SafeRLHF} \\
$\rho$ & E[c] & Drift & P(ch) & E[c] & Drift & P(ch) & E[c] & Drift & P(ch) & E[c] & Drift & P(ch) & E[c] & Drift & P(ch) & E[c] & Drift & P(ch) \\
\midrule
Base & 1227 & --- & .454 & .363 & --- & .351 & .525 & --- & .439 & .363 & --- & .420 & 126.5 & --- & .511 & .4956 & --- & .396 \\
0.0 & 1182 & $-$46 & .489 & .225 & $-$.138 & .358 & .409 & $-$.116 & .473 & .227 & $-$.136 & .516 & 121.4 & $-$5.1 & .479 & .4915 & $-$.004 & .418 \\
0.3 & 1197 & $-$30 & .532 & .379 & +.016 & .423 & .506 & $-$.019 & .514 & .337 & $-$.026 & .597 & 125.9 & $-$0.6 & .556 & .4922 & $-$.003 & .418 \\
0.5 & 1212 & $-$15 & .550 & .392 & +.029 & .434 & .519 & $-$.006 & .523 & .363 & .000 & .623 & 126.8 & +0.3 & .566 & .4923 & $-$.003 & .418 \\
0.7 & 1223 & $-$5 & .560 & .404 & +.041 & .443 & .530 & +.005 & .528 & .375 & +.012 & .631 & 127.2 & +0.7 & .568 & .4926 & $-$.003 & .417 \\
1.0 & 1247 & +20 & .566 & .422 & +.059 & .453 & .544 & +.019 & .532 & .389 & +.026 & .652 & 127.5 & +1.0 & .567 & .4928 & $-$.003 & .417 \\
\bottomrule
\end{tabular}
\end{table}

\subsection{Comparison with Prior Work}
\label{sec:exp_baselines}
Table~\ref{tab:baselines} compares CoDPO with five recent baselines across four conflict-dominated domains, and the results show three patterns.
First, SafeDPO~\citep{kim2025safedpo_iclr2026} and PD-DPO~\citep{du2025primal_arxiv2025}, which enforce a constraint on the expected cost, lower the constraint cost well below the reference level, and their $P(\mathrm{chosen})$ remains near or below that of the base model.
C2-DPO~\citep{asadi2025c2_arxiv2025}, which constrains the update itself, keeps both the constraint cost and $P(\mathrm{chosen})$ near their initial values.
These methods are designed to satisfy a constraint rather than to retain preference learning at limited drift, which is what our evaluation measures.
Second, MODPO~\citep{zhou2024beyond_acl2024} and Rewarded Soups~\citep{rame2023rewarded_nips2023} expose a trade-off hyperparameter, and sweeping it traces a frontier between drift and preference learning (Tables~6 and~7 in Appendix~D).
Third, among all methods and configurations with drift comparable to or below that of CoDPO, CoDPO attains the highest $P(\mathrm{chosen})$ in all four domains, and the configurations with higher $P(\mathrm{chosen})$, MODPO with constraint weight $w \le 0.5$, have roughly 1.5 to 5 times its drift.
Relative to vanilla DPO, CoDPO also reduces drift while raising $P(\mathrm{chosen})$ in all four domains.

\subsection{Contributions of Coarse Weights and Margin Shift}
\label{sec:exp_codpo}
Table~\ref{tab:codpo} separates the contributions of the two components by comparing vanilla DPO, coarse weights alone (Eq.~\ref{eq:w_exp} with $\lambda{=}0$), and the full CoDPO (coarse weights with the adaptive margin shift) across four conflict-dominated domains.
Coarse control alone reduces drift in all four domains.
Adding the margin shift reduces drift further in every case.
This additional reduction is largest on GSM8K ($+$17 percentage points) and smallest on HelpSteer2 ($+$2 percentage points).

\begin{table}[t]
\centering
\caption{Comparison with prior work across four domains. Drift $= \mathbb{E}[c(\theta)] - \mathbb{E}[c(\theta_\mathrm{ref})]$. P(ch) $= P_\theta(y_w \succ y_l \mid x)$. MODPO ($w{=}1.0$) and Soups ($\alpha{=}0.75$) use the setting with the smallest $|$drift$|$. Full hyperparameter sweeps are in Appendix~D. Implementation details are given in Appendix~A.}
\label{tab:baselines}
\scriptsize
\begin{tabular}{l cc cc cc cc}
\toprule
 & \multicolumn{2}{c}{HelpSteer2} & \multicolumn{2}{c}{GSM8K} & \multicolumn{2}{c}{ARC} & \multicolumn{2}{c}{Math-Calc} \\
\cmidrule(lr){2-3} \cmidrule(lr){4-5} \cmidrule(lr){6-7} \cmidrule(lr){8-9}
Method & Drift & P(ch) & Drift & P(ch) & Drift & P(ch) & Drift & P(ch) \\
\midrule
Base & --- & .454 & --- & .351 & --- & .439 & --- & .420 \\
Vanilla & $+$59 & .566 & $+$.129 & .453 & $+$.055 & .532 & $+$.071 & .652 \\
\midrule
SafeDPO & $-$132 & .437 & $-$.353 & .284 & $-$.493 & .412 & $-$.352 & .442 \\
PD-DPO & $-$128 & .452 & $-$.360 & .289 & $-$.509 & .427 & $-$.357 & .449 \\
C2-DPO & $\approx$0 & .457 & $+$.004 & .353 & $+$.003 & .442 & $\approx$0 & .423 \\
MODPO & $-$74 & .502 & $-$.180 & .365 & $-$.158 & .466 & $-$.175 & .514 \\
Soups & $+$3 & .509 & $+$.018 & .406 & $-$.010 & .480 & $+$.001 & .540 \\
\midrule
\textbf{CoDPO} & $+$18.7 & \textbf{.640} & $+$.082 & \textbf{.551} & $+$.046 & \textbf{.622} & $+$.066 & \textbf{.757} \\
\bottomrule
\end{tabular}
\end{table}

\begin{table}[t]
\centering
\caption{Drift under vanilla training, coarse control only, and full CoDPO, across domains (Qwen2.5-7B), base models (HelpSteer2), and preference losses (HelpSteer2; IPO with $\beta{=}0.5$, SimPO with $\beta{=}2$, $\gamma{=}0.5$; coarse control at $\rho^*_\mathrm{eq}$). Coarse control reduces drift in every setting.}
\label{tab:codpo}\label{tab:model_agnostic}\label{tab:algo_agnostic}
\scriptsize
\setlength{\tabcolsep}{3.4pt}
\begin{tabular}{l cccc ccc ccc}
\toprule
 & \multicolumn{4}{c}{Domains (Qwen2.5-7B)} & \multicolumn{3}{c}{Base models (HelpSteer2)} & \multicolumn{3}{c}{Losses (HelpSteer2)} \\
\cmidrule(lr){2-5}\cmidrule(lr){6-8}\cmidrule(lr){9-11}
 & HelpSteer2 & GSM8K & ARC & Math-Calc & DeepSeek-8B & Mistral-7B & Qwen3-8B & IPO & SimPO & KTO \\
\midrule
Vanilla  & $+$59   & $+$.129 & $+$.055 & $+$.071 & $+$18 & $+$92 & $+$34  & $+$15 & $+$74 & $+$72 \\
Coarse   & $+$20   & $+$.105 & $+$.049 & $+$.068 & $-$7  & $+$45 & $+$0.4 & $-$36 & $-$1  & $+$16 \\
Combined & $+$18.7 & $+$.082 & $+$.046 & $+$.066 & $-$7  & $+$41 & $+$0.4 & ---   & ---   & --- \\
\bottomrule
\end{tabular}
\end{table}

\begin{table}[t]
\centering
\caption{LLM-as-judge evaluation (JudgeLM-13B) on HelpSteer2: CoDPO compared pairwise against each method.}
\label{tab:judge}
\scriptsize
\begin{tabular}{l cccccc}
\toprule
 & Vanilla DPO & SafeDPO & PD-DPO & C2-DPO & MODPO & Soups \\
\midrule
CoDPO wins & 14.1\% & 86.3\% & 73.9\% & 12.7\% & 52.2\% & 13.5\% \\
Opp.\ wins & 10.5\% & 0.0\%  & 3.0\%  & 10.7\% & 7.7\%  & 10.0\% \\
Tie        & 75.5\% & 13.6\% & 23.2\% & 76.6\% & 40.1\% & 76.5\% \\
\bottomrule
\end{tabular}
\end{table}

\subsection{Robustness Across Models and Losses}
\label{sec:exp_generalization}
On HelpSteer2, across four base models (Qwen2.5-7B~\citep{yang2024qwen25_arxiv2024}, DeepSeek-8B~\citep{guo2025deepseek_arxiv2025}, Mistral-7B~\citep{jiang2023mistral_arxiv2023}, Qwen3-8B~\citep{yang2025qwen3_arxiv2025}; Table~\ref{tab:model_agnostic}), vanilla DPO drifts upward in every case and the coarse weights reduce drift in every case, with the margin shift reducing it further on Qwen2.5 and Mistral.
Applied as a data-level plug-in to IPO~\citep{azar2024general_aistas2024}, SimPO~\citep{meng2024simpo_nips2024}, and KTO~\citep{ethayarajh2024kto_arxiv2024} (Table~\ref{tab:algo_agnostic}), the coarse weights also reduce drift in all three cases.

\subsection{LLM-as-Judge Evaluation}
\label{sec:exp_judge}
To evaluate generation quality beyond probability-level metrics, we compare CoDPO pairwise with each method on all 7{,}220 HelpSteer2 prompts using JudgeLM-13B~\citep{zhu2023judgelm_iclr2025} with position-debiased scoring (Table~\ref{tab:judge}), and CoDPO wins more often than it loses against every compared method.
Compared with SafeDPO~\citep{kim2025safedpo_iclr2026}, PD-DPO~\citep{du2025primal_arxiv2025}, and MODPO~\citep{zhou2024beyond_acl2024}, CoDPO wins 52\%--86\% of comparisons, in line with their lower $P(\mathrm{chosen})$ in Table~\ref{tab:baselines}.
Compared with vanilla DPO~\citep{rafailov2023direct_nips2023}, C2-DPO~\citep{asadi2025c2_arxiv2025}, and Rewarded Soups~\citep{rame2023rewarded_nips2023}, most comparisons are ties (75\%--77\%), so CoDPO reduces drift relative to vanilla DPO without a change in generation quality detectable by this judge.
GPT-4o as a second judge agrees: CoDPO ties vanilla DPO and is preferred over MODPO 44.9\% to 3.9\% (Appendix~F.3).

\subsection{Ablations}
\label{sec:exp_ablations}
\noindent\textbf{Steepness parameter $p$.}
The weighting family of Eq.~\ref{eq:w_exp} recovers the PSDPO weighting at $p{=}1$ and approaches a hard threshold on conflict severity as $p{\to}\infty$.
Table~8 sweeps $p$ on HelpSteer2: drift reduction increases monotonically with $p$ (22\% to 75\%), while $P(\mathrm{chosen})$ remains essentially constant (.640--.644).
Due to compute constraints, we use $p{=}8$ in all experiments without tuning it per domain, although larger $p$ lowers drift further here.

\noindent\textbf{Dynamic diagnostic validity.}
To examine the initialization approximation $\sigma(\beta\hat{r}_\theta) \approx 1/2$, we track $\rho^*(t)$ during vanilla DPO training on HelpSteer2, recomputing only the learning signal from the actual $\sigma(\beta\hat{r}_\theta)$ on all 7{,}220 pairs: $\rho^*(0) = 0.710$ matches $\rho^*_\mathrm{eq} = 0.709$, and as $\sigma \to 0.946$, $\rho^*(t)$ moves to 0.603 (gap 0.11).
This is consistent with the balance point shifting during training, which motivates the adaptive margin shift; Appendix~E reports the trajectory of $\lambda$ during training.

\section{Conclusion}
\label{sec:conclusion}
We studied constraint drift under heterogeneous supervision and showed that, for small updates, it depends on the balance between conflict-side pressure and matching-side repair.
From this balance we derive a diagnostic, $\rho^*_\mathrm{eq}$, that is computed from dataset statistics before training and indicates whether a dataset needs intervention.
When it does, CoDPO down-weights conflict pairs through the asymmetric weighting of PSDPO~\citep{meng2026psdpo_mm2026}, with a strength set by $\rho^*_\mathrm{eq}$ rather than tuned, and shifts the DPO margin during training to address the remaining drift.
Empirically, the intervention decision given by $\rho^*_\mathrm{eq}$ matches the observed drift on all evaluated domains, and in our main comparison CoDPO reduces drift while largely preserving the preference learning of vanilla DPO.
A natural next step is to track the balance as the model changes during training, rather than only at initialization, and to extend the analysis to settings where the preference pairs themselves are generated during training~\citep{xu2026gpdpo_arxiv2026}.

\begin{ack}
This work used the high-performance GPU cluster acquired with support from the U.S. National Science Foundation (NSF) under Award \#2511989, ``Research Infrastructure: MRI: Acquisition of a High-Performance GPU Cluster for Large-Scale AI.''
This work also used Delta and DeltaAI at the National Center for Supercomputing Applications (NCSA) through allocation CIS260780 from the Advanced Cyberinfrastructure Coordination Ecosystem: Services \& Support (ACCESS) program~\citep{boerner2023access}, which is supported by NSF grants \#2138259, \#2138286, \#2138307, \#2137603, and \#2138296.
Any opinions, findings, and conclusions or recommendations expressed in this material are those of the authors and do not necessarily reflect the views of the NSF.
The authors declare no competing interests.
\end{ack}

{\small
\bibliographystyle{unsrt}
\bibliography{references}

\begin{thebibliography}{10}

\bibitem{stiennon2020learning_nips2020}
Nisan Stiennon, Long Ouyang, Jeffrey Wu, Daniel Ziegler, Ryan Lowe, Chelsea
  Voss, Alec Radford, Dario Amodei, and Paul~F Christiano.
\newblock Learning to summarize with human feedback.
\newblock In {\em NeurIPS}, 2020.

\bibitem{ouyang2022training_nips2022}
Long Ouyang, Jeffrey Wu, Xu~Jiang, Diogo Almeida, Carroll Wainwright, Pamela
  Mishkin, Chong Zhang, Sandhini Agarwal, Katarina Slama, Alex Ray, et~al.
\newblock Training language models to follow instructions with human feedback.
\newblock In {\em NeurIPS}, 2022.

\bibitem{bai2022training_arxiv2022}
Yuntao Bai, Andy Jones, Kamal Ndousse, Amanda Askell, Anna Chen, Nova DasSarma,
  Dawn Drain, Stanislav Fort, Deep Ganguli, Tom Henighan, et~al.
\newblock Training a helpful and harmless assistant with reinforcement learning
  from human feedback.
\newblock {\em arXiv preprint arXiv:2204.05862}, 2022.

\bibitem{rafailov2023direct_nips2023}
Rafael Rafailov, Archit Sharma, Eric Mitchell, Christopher~D Manning, Stefano
  Ermon, and Chelsea Finn.
\newblock Direct preference optimization: Your language model is secretly a
  reward model.
\newblock In {\em NeurIPS}, 2023.

\bibitem{azar2024general_aistas2024}
Mohammad Gheshlaghi~Azar, Zhaohan~Daniel Guo, Bilal Piot, R{\'e}mi Munos, Mark
  Rowland, Michal Valko, and Daniele Calandriello.
\newblock A general theoretical paradigm to understand learning from human
  preferences.
\newblock In {\em AISTATS}, 2024.

\bibitem{ethayarajh2024kto_arxiv2024}
Kawin Ethayarajh, Winnie Xu, Niklas Muennighoff, Dan Jurafsky, and Douwe Kiela.
\newblock Model alignment as prospect theoretic optimization.
\newblock In {\em ICML}, 2024.

\bibitem{meng2024simpo_nips2024}
Yu~Meng, Mengzhou Xia, and Danqi Chen.
\newblock Sim{PO}: Simple preference optimization with a reference-free reward.
\newblock In {\em NeurIPS}, 2024.

\bibitem{guo2025deepseek_arxiv2025}
Daya Guo, Dejian Yang, Haowei Zhang, et~al.
\newblock {DeepSeek-R1} incentivizes reasoning in {LLMs} through reinforcement
  learning.
\newblock {\em Nature}, 645:633--638, 2025.

\bibitem{dai2023safe_arxiv2023}
Juntao Dai, Xuehai Pan, Ruiyang Sun, Jiaming Ji, Xinbo Xu, Mickel Liu, Yizhou
  Wang, and Yaodong Yang.
\newblock Safe {RLHF}: Safe reinforcement learning from human feedback.
\newblock In {\em ICLR}, 2024.

\bibitem{ji2025pku_acl2025}
Jiaming Ji, Donghai Hong, Borong Zhang, Boyuan Chen, Josef Dai, Boren Zheng,
  Tianyi~Alex Qiu, Jiayi Zhou, Kaile Wang, Boxun Li, et~al.
\newblock {PKU-SafeRLHF}: Towards multi-level safety alignment for {LLMs} with
  human preference.
\newblock In {\em ACL}, 2025.

\bibitem{meng2026psdpo_mm2026}
Siwei Meng, Yawei Luo, Shu Zhang, and Ping Liu.
\newblock When physical preferences meet semantic constraints: Physical and
  semantic direct preference optimization for text-to-video generation.
\newblock In {\em ACM MM}, 2026.

\bibitem{wang2024helpsteer_nips2024}
Zhilin Wang, Yi~Dong, Olivier Delalleau, Jiaqi Zeng, Gerald Shen, Daniel Egert,
  Jimmy~J Zhang, Makesh~Narsimhan Sreedhar, and Oleksii Kuchaiev.
\newblock {HelpSteer} 2: Open-source dataset for training top-performing reward
  models.
\newblock In {\em NeurIPS Datasets and Benchmarks Track}, 2024.

\bibitem{cobbe2021training_arxiv2021}
Karl Cobbe, Vineet Kosaraju, Mohammad Bavarian, Mark Chen, Heewoo Jun, Lukasz
  Kaiser, Matthias Plappert, Jerry Tworek, Jacob Hilton, Reiichiro Nakano,
  et~al.
\newblock Training verifiers to solve math word problems.
\newblock {\em arXiv preprint arXiv:2110.14168}, 2021.

\bibitem{clark2018think_arxiv2018}
Peter Clark, Isaac Cowhey, Oren Etzioni, Tushar Khot, Ashish Sabharwal, Carissa
  Schoenick, and Oyvind Tafjord.
\newblock Think you have solved question answering? {T}ry {ARC}, the {AI2}
  reasoning challenge.
\newblock {\em arXiv preprint arXiv:1803.05457}, 2018.

\bibitem{wachi2024stepwise_nips2024}
Akifumi Wachi, Thien~Q Tran, Rei Sato, Takumi Tanabe, and Youhei Akimoto.
\newblock Stepwise alignment for constrained language model policy
  optimization.
\newblock In {\em NeurIPS}, 2024.

\bibitem{du2025primal_arxiv2025}
Yihan Du, Seo~Taek Kong, and R.~Srikant.
\newblock Provably convergent primal-dual {DPO} for constrained {LLM}
  alignment.
\newblock {\em arXiv preprint arXiv:2510.05703}, 2025.

\bibitem{asadi2025c2_arxiv2025}
Kavosh Asadi, Xingzi Xu, Julien Han, Ege Beyazit, Idan Pipano, Dominique
  Perrault-Joncas, Shoham Sabach, Mohammad Ghavamzadeh, and Karim Bouyarmane.
\newblock {C2-DPO}: Constrained controlled direct preference optimization.
\newblock {\em TMLR}, 2026.

\bibitem{kim2025safedpo_iclr2026}
Geon-Hyeong Kim, Yu~Jin Kim, Byoungjip Kim, Honglak Lee, Kyunghoon Bae,
  Youngsoo Jang, and Moontae Lee.
\newblock {SafeDPO}: A simple approach to direct preference optimization with
  enhanced safety.
\newblock In {\em ICLR}, 2026.

\bibitem{rame2023rewarded_nips2023}
Alexandre Rame, Guillaume Couairon, Corentin Dancette, Jean-Baptiste Gaya,
  Mustafa Shukor, Laure Soulier, and Matthieu Cord.
\newblock Rewarded soups: towards {P}areto-optimal alignment by interpolating
  weights fine-tuned on diverse rewards.
\newblock In {\em NeurIPS}, 2023.

\bibitem{zhou2024beyond_acl2024}
Zhanhui Zhou, Jie Liu, Jing Shao, Xiangyu Yue, Chao Yang, Wanli Ouyang, and
  Yu~Qiao.
\newblock Beyond one-preference-fits-all alignment: Multi-objective direct
  preference optimization.
\newblock In {\em Findings of ACL}, 2024.

\bibitem{deng2025less_nips2025}
Xun Deng, Han Zhong, Rui Ai, Fuli Feng, Zheng Wang, and Xiangnan He.
\newblock Less is more: Improving {LLM} alignment via preference data
  selection.
\newblock In {\em NeurIPS}, 2025.

\bibitem{penguni_iclr2026}
Shangpin Peng, Weinong Wang, Zhuotao Tian, Senqiao Yang, Xing Wu, Haotian Xu,
  Chengquan Zhang, Takashi Isobe, Baotian Hu, and Min Zhang.
\newblock {Uni-DPO}: A unified paradigm for dynamic preference optimization of
  {LLMs}.
\newblock In {\em ICLR}, 2026.

\bibitem{xiao2024cal_nips2024}
Teng Xiao, Yige Yuan, Huaisheng Zhu, Mingxiao Li, and Vasant~G Honavar.
\newblock {Cal-DPO}: Calibrated direct preference optimization for language
  model alignment.
\newblock In {\em NeurIPS}, 2024.

\bibitem{chowdhury2024provably_icml2024}
Sayak Ray~Chowdhury, Anush Kini, and Nagarajan Natarajan.
\newblock Provably robust {DPO}: Aligning language models with noisy feedback.
\newblock In {\em ICML}, 2024.

\bibitem{wu2024beta_nips2024}
Junkang Wu, Yuexiang Xie, Zhengyi Yang, Jiancan Wu, Jinyang Gao, Bolin Ding,
  Xiang Wang, and Xiangnan He.
\newblock $\beta$-{DPO}: Direct preference optimization with dynamic $\beta$.
\newblock In {\em NeurIPS}, 2024.

\bibitem{li2025self_acl2025}
Moxin Li, Yuantao Zhang, Wenjie Wang, Wentao Shi, Zhuo Liu, Fuli Feng, and
  Tat-Seng Chua.
\newblock Self-improvement towards {P}areto optimality: Mitigating preference
  conflicts in multi-objective alignment.
\newblock In {\em Findings of ACL}, 2025.

\bibitem{xu2026understanding_acl2026}
Zhihao Xu, Yongqi Tong, Xin Zhang, Jun Zhou, and Xiting Wang.
\newblock Understanding conflicts in multi-objective alignment through reward
  consistency.
\newblock In {\em Findings of ACL}, 2026.

\bibitem{zhangdata_iclr2026}
Jia Zhang, Yao Liu, Chen-Xi Zhang, Yi~Liu, Yi-Xuan Jin, Lan-Zhe Guo, and
  Yu-Feng Li.
\newblock Data selection for {LLM} alignment using fine-grained preferences.
\newblock In {\em ICLR}, 2026.

\bibitem{chen2026reward_icml2026}
Peter Chen, Xiaopeng Li, Xi~Chen, and Tianyi Lin.
\newblock Reward-free alignment for conflicting objectives.
\newblock In {\em ICML}, 2026.

\bibitem{zhang2025bifactorial_iclr2025}
Wenxuan Zhang, Philip H.~S. Torr, Mohamed Elhoseiny, and Adel Bibi.
\newblock Bi-factorial preference optimization: Balancing safety-helpfulness in
  language models.
\newblock In {\em ICLR}, 2025.

\bibitem{yu2020gradient_nips2020}
Tianhe Yu, Saurabh Kumar, Abhishek Gupta, Sergey Levine, Karol Hausman, and
  Chelsea Finn.
\newblock Gradient surgery for multi-task learning.
\newblock In {\em NeurIPS}, 2020.

\bibitem{liu2021conflict_nips2021}
Bo~Liu, Xingchao Liu, Xiaojie Jin, Peter Stone, and Qiang Liu.
\newblock Conflict-averse gradient descent for multi-task learning.
\newblock In {\em NeurIPS}, 2021.

\bibitem{yang2025selective_emnlp2025}
Kailai Yang, Zhiwei Liu, Qianqian Xie, Jimin Huang, Erxue Min, and Sophia
  Ananiadou.
\newblock Selective preference optimization via token-level reward function
  estimation.
\newblock In {\em EMNLP}, 2025.

\bibitem{singhal2024long_colm2024}
Prasann Singhal, Tanya Goyal, Jiacheng Xu, and Greg Durrett.
\newblock A long way to go: Investigating length correlations in {RLHF}.
\newblock In {\em COLM}, 2024.

\bibitem{park2024disentangling_acl2024}
Ryan Park, Rafael Rafailov, Stefano Ermon, and Chelsea Finn.
\newblock Disentangling length from quality in direct preference optimization.
\newblock In {\em Findings of ACL}, 2024.

\bibitem{razin2025unintentional_iclr2025}
Noam Razin, Sadhika Malladi, Adithya Bhaskar, Danqi Chen, Sanjeev Arora, and
  Boris Hanin.
\newblock Unintentional unalignment: Likelihood displacement in direct
  preference optimization.
\newblock In {\em ICLR}, 2025.

\bibitem{ren2025learning_iclr2025}
Yi~Ren and Danica~J. Sutherland.
\newblock Learning dynamics of {LLM} finetuning.
\newblock In {\em ICLR}, 2025.

\bibitem{amini2024odpo_acl2024}
Afra Amini, Tim Vieira, and Ryan Cotterell.
\newblock Direct preference optimization with an offset.
\newblock In {\em Findings of ACL}, 2024.

\bibitem{yang2024qwen25_arxiv2024}
{Qwen}, An~Yang, Baosong Yang, Beichen Zhang, Binyuan Hui, et~al.
\newblock {Qwen2.5} technical report.
\newblock {\em arXiv preprint arXiv:2412.15115}, 2024.

\bibitem{hu2022lora_iclr2022}
Edward~J. Hu, Yelong Shen, Phillip Wallis, Zeyuan Allen-Zhu, Yuanzhi Li, Shean
  Wang, Lu~Wang, and Weizhu Chen.
\newblock {LoRA}: Low-rank adaptation of large language models.
\newblock In {\em ICLR}, 2022.

\bibitem{jiang2023mistral_arxiv2023}
Albert~Q. Jiang, Alexandre Sablayrolles, Arthur Mensch, et~al.
\newblock {Mistral 7B}.
\newblock {\em arXiv preprint arXiv:2310.06825}, 2023.

\bibitem{yang2025qwen3_arxiv2025}
An~Yang et~al.
\newblock Qwen3 technical report.
\newblock {\em arXiv preprint arXiv:2505.09388}, 2025.

\bibitem{zhu2023judgelm_iclr2025}
Lianghui Zhu, Xinggang Wang, and Xinlong Wang.
\newblock {JudgeLM}: Fine-tuned large language models are scalable judges.
\newblock In {\em ICLR}, 2025.

\bibitem{xu2026gpdpo_arxiv2026}
Zixuan Xu, Haoxuan Chen, Jiangxin Sun, and Jian-Fang Hu.
\newblock {GPDPO}: Learning from reward conflicts via group {P}areto-dominance
  policy optimization.
\newblock {\em arXiv preprint arXiv:2610.35480}, 2026.

\bibitem{boerner2023access}
Timothy~J. Boerner, Stephen Deems, Thomas~R. Furlani, Shelley~L. Knuth, and
  John Towns.
\newblock {ACCESS}: Advancing innovation: {NSF}'s advanced cyberinfrastructure
  coordination ecosystem: Services \& support.
\newblock In {\em Practice and Experience in Advanced Research Computing (PEARC
  '23)}, 2023.

\end{thebibliography}
}

\end{document}